\documentclass[preprint,12pt]{elsarticle}

\usepackage{amssymb}
\usepackage{amsmath}
\journal{Neurocomputing}

\begin{document}

\begin{frontmatter}



\title{High-Performance Inference Graph Convolutional Networks for Skeleton-Based Action Recognition}

\author[1,3]{Junyi Wang}
\ead{wjyi168@126.com}

\author[1,2]{Ziao Li \corref{cor}}
\ead{liziao26@163.com} 

\author[4]{Bangli Liu}
\ead{jlliubangli@gmail.com}

\author[5]{Haibin Cai}
\ead{H.Cai@lboro.ac.uk} 

\author[5]{Mohamad Saada}
\ead{M.Saada@lboro.ac.uk} 

\author[5]{Qinggang Meng}
\ead{Q.Meng@lboro.ac.uk}

\affiliation[1]{organization={Robot Science and Engineering},%Department and Organization
            addressline={Northeastern University}, 
            city={Shenyang},
            postcode={110819}, 
            state={Liaoning},
            country={China}}

\affiliation[2]{organization={School of Intelligent Systems Engineering},
            addressline={Sun Yat-sen University},
            city={Shenzhen},
            postcode={518107},
            state={Guangdong},
            country={China}}

\affiliation[3]{organization={Foshan Graduate School of Innovation},
            addressline={Northeastern University},
            city={Foshan},
            postcode={528311},
            state={Guangdong},
            country={China}}

\affiliation[4]{organization={Computer Science and Informatics},
            addressline={De Montfort University},
            city={Leicester},
            postcode={LE1 9BH},
            country={United Kingdom}}

\affiliation[5]{organization={Department of Computer Science},
            addressline={Loughborough University},
            city={Loughborough},
            postcode={LE11 3TU},
            country={United Kingdom}}

\cortext[cor]{Corresponding author}

%% Abstract
\begin{abstract}
Recently, the significant achievements have been made in skeleton-based human action recognition with the emergence of graph convolutional networks (GCNs). However, the state-of-the-art (SOTA) models used for this task focus on constructing more complex higher-order connections between joint nodes to describe skeleton information, which leads to complex inference processes and high computational costs. To address the slow inference speed caused by overly complex model structures, we introduce re-parameterization and over-parameterization techniques to GCNs and propose two novel high-performance inference GCNs, namely HPI-GCN-RP and HPI-GCN-OP. 
After the completion of model training, model parameters are fixed. HPI-GCN-RP adopts re-parameterization technique to transform high-performance training model into fast inference model through linear transformations, which achieves a higher inference speed with competitive model performance. 
HPI-GCN-OP further utilizes over-parameterization technique to achieve higher performance improvement by introducing additional inference parameters, albeit with slightly decreased inference speed.
The experimental results on the two skeleton-based action recognition datasets demonstrate the effectiveness of our approach. 
Our HPI-GCN-OP achieves performance comparable to the current state-of-the-art (SOTA) models, with inference speeds five times faster. Specifically, our HPI-GCN-OP achieves an accuracy of 93\% on the cross-subject split of the NTU-RGB+D 60 dataset, and 90.1\% on the cross-subject benchmark of the NTU-RGB+D 120 dataset. Code is available at github.com/lizaowo/HPI-GCN.
\end{abstract}

%\includegraphics[width=1.1\textwidth]{g-abs.eps}

%	\item In skeleton-based action recognition, we are the first to introduce re-parameterization into GCN and design HPI-GCN-RP. By employing a multi-branch training structure and a simple inference structure, HPI-GCN-RP achieves 1.5 times faster inference speed and 2.1\% performance improvement compared to ST-GCN on the NTU-RGB+D 120 dataset.

%	\item We further introduce the over-parameterization to HPI-GCN-RP and propose HPI-GCN-OP. By augmenting the non-linear learnable adjacency matrix to enhance the model's representational capacity. On the NTU-RGB+D 120 and NTU-RGB+D 60 datasets, HPI-GCN-OP achieves the comparable performance to the current SOTA while boasting an inference speed five times faster.

%	\item Our re-parameterization method demonstrates strong generalization. By employing our Rep-TCN to replace the temporal convolution block in current models, it simultaneously enhances both model performance and inference speed.

%% Keywords

\end{frontmatter}


\section{Introduction}\label{sec1}
In the context of deep learning, the human action recognition has been applied to various tasks \cite{MADT-GCN, YUE2022287}, such as motion analysis, sign language translation, and human-computer interaction. With the advent of ST-GCN \cite{stgcn}, the graph convolutional method has become a general approach for skeleton-based human action recognition. By utilizing a multi-branch learnable topology of human skeleton, the relationship between joints can be represented simply and efficiently. This method has the advantages of simple structure and fast inference speed, but the accuracy on the NTU-RGB+D 120 dataset is only 83.4\%.

\begin{figure}
  \centering
  \setlength{\abovecaptionskip}{0.1cm}
  \includegraphics[width=0.8\linewidth]{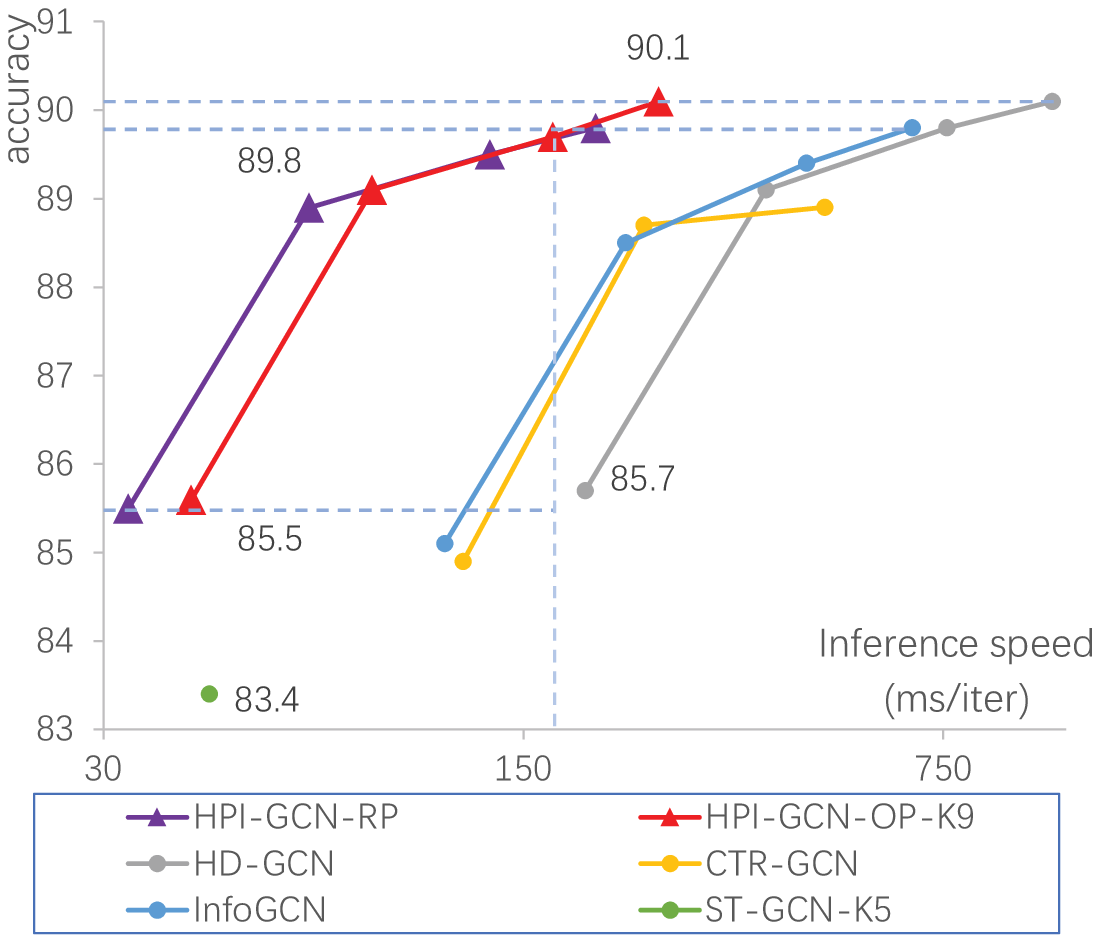}
  \caption{Comparison of inference speeds for skeleton-based action analysis. The computational speed on an NVIDIA RTX 3090 GPU with a batch size of 64 and full precision (fp32), which is measured in milliseconds / iteration. The x-axis is logarithmic with a base of 10. The points of different positions in the same color represent models with different number of streams. The dataset is NTU-RGB+D 120 X-sub. K9 stands for time convolution with a kernel of 9$\times$1. Triangles represent our proposed HPI-GCN, circles represent mainstream methods. It can be seen that our performance curve is better than SOTA models. Please refer to Tab. \ref{Tab1} for more details.}
  \label{F1}
\end{figure}

Recent works\cite{asgcn, 2sagcn, song2022constructing, stgcnplus, msg3d, ctrgcn, infogcn, hdgcn} have shown improvements in accuracy compared to ST-GCN. ST-GCN++ \cite{stgcnplus} replaced temporal convolution networks (TCN) with multi-scale temporal convolution networks (MS-TCN) \cite{msg3d} and adopted well-designed training strategies, resulting in a significant accuracy boost for ST-GCN. CTR-GCN \cite{ctrgcn} collected behavior context information from skeleton data to model the intrinsic topology between joints. InfoGCN \cite{infogcn} utilized attention-based graph convolution to capture the intrinsic topology of human action. HD-GCN \cite{hdgcn} proposed hierarchically decomposed GCN to model the graph topology. However, these works focus on constructing more complex higher-order connections between skeleton data to improve model performance, paying less attention to real-time performance.
These multi-branch structures and higher-order connections introduces non-contiguous memory access, which is unfriendly for strong parallel computing powers like GPU \cite{shufflenetv2}. In reality, although current SOTA models achieve commendable accuracy, their inference speed is suboptimal, making them impractical and costly for widespread industrial applications. Consequently, there is a pressing need to design a model that strikes a harmonious balance between inference speed and performance, addressing the practical limitations posed by existing models.

The structural reparameterization technique has demonstrated excellent performance in lightweight convolutional neural networks \cite{rep0,rep1,acnet,repvgg, dbb}, obtaining a simplified inference model by applying linear transformations to the training model. This method reduces the model's parameter count, accelerates inference speed, while maintaining model performance. However, skeleton point behavior analysis typically involves processing time series data, where the connectivity of skeleton points changes over time. Directly applying re-parameterization methods from convolutional neural networks is not feasible. We reorganize and redesign the structural re-parameterization technique and graph convolution to adapt to skeleton data, addressing challenges in current skeleton-based action analysis tasks.

For the spatial features of skeleton-based action analysis, we propose a high-performance inference graph convolution (HPI-GC-RP). It is based on the proposed Adjacency matrix multiplication fusion, a re-parameterization technique for fixed-number node graph convolution. Specifically, during training, parallel multiple adjacency matrices are performed to extract skeleton information, and during inference, multiple adjacency matrices are fused into a single one to reduce model complexity. For temporal features, we propose re-parameterized temporal convolution (Rep-TCN), which utilizes a multi-branch structure for training to extract multi-scale temporal information. Through re-parameterization techniques, it linearly transforms the multi-branch structure into a single-branch inference structure, significantly accelerating the model's inference speed while preserving its performance. Through ablation experiments, our Rep-TCN directly replaces the temporal convolution block in current models, it simultaneously enhances both model performance and inference speed.
Based on Adjacency matrix multiplication fusion and Rep-TCN, we enhance proposed HPI-GCN-RP. This model simultaneously considers performance and inference speed, addressing the difficulties associated with reparameterization in the context of graph convolution for time-varying skeletal data connections.

As shown in Fig. \ref{F1}, on the NTU-RGB+D 120 cross-subject (X-sub), our single-stream HPI-GCN-RP has a 1.5 times faster inference speed than ST-GCN and 2.1\% higher accuracy. Compared with InfoGCN \cite{infogcn}, our HPI-GCN-RP achieves better precision on single-stream, identical precision on six-stream, and the inference speed is 3.4 times faster, which shows that our HPI-GCN-RP has achieved a better balance between inference speed and accuracy.
While the re-parameterization enhances the inference speed, it does not improve the model's expressive power. We further over-parameterizes HPI-GCN-RP by augmenting the non-linear learnable adjacency matrix to enhance the model's representational capacity. Additionally, we design a more powerful module named HPI-GCN-OP. As shown in Fig. \ref{F1}, there is a significant improvement in accuracy with a marginal increase in inference time. The six-stream HPI-GCN-OP-K9 has the same accuracy as the current SOTA model HD-GCN \cite{hdgcn}, but our inference speed is 4.5 times faster than HD-GCN. When inferring a batch size of 64 per iteration, the inference speed of four-stream HPI-GCN-OP-K9 is 23ms faster than the single-stream HD-GCN with an accuracy of 89.7\% that is 4\% higher than HD-GCN.

Our main contributions are as follows:
\begin{itemize}
	\item In skeleton-based action recognition, we are the first to introduce re-parameterization into GCN and design HPI-GCN-RP. By employing a multi-branch training structure and a simple inference structure, HPI-GCN-RP achieves 1.5 times faster inference speed and 2.1\% performance improvement compared to ST-GCN on the NTU-RGB+D 120 dataset.

	\item We further introduce the over-parameterization to HPI-GCN-RP and propose HPI-GCN-OP. By augmenting the non-linear learnable adjacency matrix to enhance the model's representational capacity. On the NTU-RGB+D 120 and NTU-RGB+D 60 datasets, HPI-GCN-OP achieves the comparable performance to the current SOTA while boasting an inference speed five times faster.

	\item Our re-parameterization method demonstrates strong generalization. By employing our Rep-TCN to replace the temporal convolution block in current models, it simultaneously enhances both model performance and inference speed.
\end{itemize}
\section{Proposed Algorithm}\label{sec2}
In this section, we first introduce the re-parameterization technique, which adopts re-parameterized convolution operations to improve model accuracy without impacting the inference speed. Then, we introduce the overall architecture of the proposed HPI-GCN. Finally, we introduce our proposed Rep-TCN and two high-performance inference graph convolution (HPI-GC) modules.

%-------------------------------------------------------------------------
\subsection{Preliminaries}
In skeleton-based human action analysis, the input is $\boldsymbol X \in \mathbb{R}^{C \times T \times V}$, and the output is $\boldsymbol Y \in \mathbb{R}^{C \times T \times V}$, where $C$ represent the number of channels, $T$ represents the length of the time series, and $V$ represents the number of joints. $\boldsymbol A \in \mathbb{R}^{V \times V}$ is the adjacency matrix. The output of GCN are as follows.
\begin{equation}
  \boldsymbol Y = \boldsymbol A \boldsymbol X.
\end{equation}

The structural reparameterization we employ through the homogeneity and additivity of convolution, as follows.
\begin{equation}
  \boldsymbol Y = (p\boldsymbol W )\boldsymbol X = p(\boldsymbol W \boldsymbol X),
\label{eq3}
\end{equation}
\begin{equation}
  \boldsymbol Y = \boldsymbol W_1\boldsymbol X + \boldsymbol W_2\boldsymbol X = (\boldsymbol W_1 + \boldsymbol W_2)\boldsymbol X,
\label{eq4}
\end{equation}
where $p$ is a constant. The additivity holds only if $\boldsymbol W_1$ and $\boldsymbol W_2$ have the same configuration (e.g. number of channels, kernel size, stride, padding, etc).

%-------------------------------------------------------------------------
\subsection{Re-parameterization methods}
\begin{figure}
  \centering
  \setlength{\abovecaptionskip}{0.1cm}
  \includegraphics[width=0.6\linewidth]{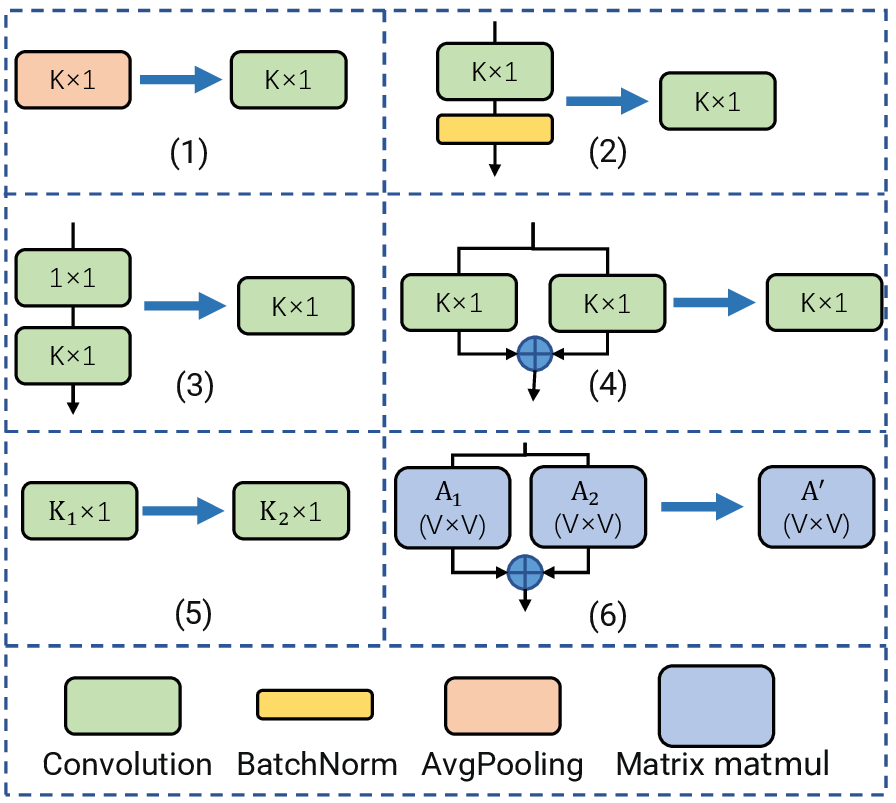}
  \caption{The six re-parameterization methods. The left side is the training mode, and the right side is the inference mode. K represents the convolution kernel, and A represents the learnable adjacency matrix.}
  \label{F2}
\end{figure}
In this section, we introduce six re-parameterization methods used in the following GCN model design. As shown in Fig. \ref{F2}, the left side is the training structure, and the right side is the inference structure. After the training is completed, the parameters are offline. Therefore, we adopt linear transformations to transform the training model into the inference model without affecting performance.
Blending 6 is the Adjacency matrix multiplication fusion, a re-parameterization method designed by us for GCN.
%which could transform average pooling, small convolution kernels, batch normalization, serial connections and parallel connections into convolutions and merge the adjacency matrix.

\textbf{Blending 1:} K $\times$ 1 average pooling to K $\times$ 1 conv, where K $\times$ 1 average pooling can be converted to K $\times$ 1 depth-wise convolution with each element of the convolution kernel as $\frac{1}{K}$ and bias $=0$. We expand it to a regular K $\times$ 1 convolution with zero padding.

\textbf{Blending 2:} K $\times$ 1 conv-bn to K $\times$ 1 conv, where the conv-bn output is given by
\begin{equation}
  \boldsymbol Y = (\boldsymbol W_1\boldsymbol X - \mu) \frac{w_{bn}}{\sigma} + b_{bn},
\end{equation}
where $\mu$ and $\sigma$ are the mean and standard deviation. $w_{bn}$ and $b_{bn}$ are the learnable scaling factor and bias of the batch norm, respectively. As the parameters of the batch normalization layer are offline in the inference process, $\frac{w_{bn}}{\sigma}$ is a constant. According to equation (\ref{eq3}), the conv-bn layer can be represented in inference as
\begin{equation}
  \boldsymbol Y = (\boldsymbol W_1 \frac{w_{bn}}{\sigma})\boldsymbol X - \mu \frac{w_{bn}}{\sigma} + b_{bn} = \boldsymbol W'\boldsymbol X + \boldsymbol B',
\end{equation}
which can be seen as a K $\times$ 1 convolution with $\boldsymbol W' = \boldsymbol W_1 \frac{w_{bn}}{\sigma}$ and $\boldsymbol B' = -\mu \frac{w_{bn}}{\sigma} + b_{bn}$.

\textbf{Blending 3:} 1 $\times$ 1 conv serializes with K $\times$ 1 conv to K $\times$ 1 conv, where the kernel structure of the 1 $\times$ 1 conv is $\boldsymbol W_1 \in \mathbb{R}^{C \times C \times 1 \times 1}$, and the kernel structure of the K $\times$ 1 conv is $\boldsymbol W_2 \in \mathbb{R}^{C \times C \times K \times 1}$. The output of the serialized layer is given by
\begin{equation}
\begin{aligned}
  \boldsymbol Y &= \boldsymbol W_2(\boldsymbol W_1\boldsymbol X + \boldsymbol B_1) + \boldsymbol B_2 \\&= \boldsymbol W_2\boldsymbol W_1\boldsymbol X + \boldsymbol W_2\boldsymbol B_1 + \boldsymbol B_2,
\label{eq7}
\end{aligned}
\end{equation}
because $\boldsymbol W_1$ only involves channel-wise interactions without spatial aggregation. It is a linear transformation of $\boldsymbol W_2$ by $\boldsymbol W_1$, so $\boldsymbol W_2\boldsymbol W_1$ can be interpreted as the process of convolving the input $\boldsymbol W_2$ with the kernel $\boldsymbol W_1$ and the output as $\boldsymbol W_1\boldsymbol W_2 \in \mathbb{R}^{C \times C \times K \times 1}$. The convolution of a constant matrix by $\boldsymbol W_2$ gives a constant matrix, and the equation (\ref{eq7}) can be written as
\begin{equation}
  \boldsymbol Y = (\boldsymbol W_1\boldsymbol W_2)\boldsymbol X + \boldsymbol W_2\boldsymbol B_1 + \boldsymbol B_2 = \boldsymbol W'\boldsymbol X + \boldsymbol B',
\end{equation}
which means that the inference model of the 1 $\times$ 1 conv serialized with a K $\times$ 1 conv can be converted to a K $\times$ 1 convolution with kernel $\boldsymbol W' = \boldsymbol W_1\boldsymbol W_2$ and bias $\boldsymbol B' = \boldsymbol W_2\boldsymbol B_1 + \boldsymbol B_2$.

\textbf{Blending 4:} K $\times$ 1 conv branch adds to K $\times$ 1 conv, where two convolutions have the same configuration, and the output is given by
\begin{equation}
\begin{aligned}
  \boldsymbol Y &= (\boldsymbol W_1\boldsymbol X + \boldsymbol B_1) + (\boldsymbol W_2\boldsymbol X + \boldsymbol B_2) \\& =  \boldsymbol W_1\boldsymbol X + \boldsymbol W_2\boldsymbol X + \boldsymbol B_1 + \boldsymbol B_2.
\end{aligned}
\end{equation}
According to equation (\ref{eq4}),
\begin{equation}
  \boldsymbol Y = (\boldsymbol W_1 + \boldsymbol W_2)\boldsymbol X + (\boldsymbol B_1 + \boldsymbol B_2) = \boldsymbol W'\boldsymbol X + \boldsymbol B',
\end{equation}
means that K $\times$ 1 conv branch added K $\times$ 1 conv can be converted to K $\times$ 1 convolution with kernel $\boldsymbol W' = \boldsymbol W_1 + \boldsymbol W_2$ and bias $\boldsymbol B' = \boldsymbol B_1 + \boldsymbol B_2$.

\textbf{Blending 5:} K$_1$ $\times$ 1 conv to K$_2$ $\times$ 1 conv, where $K_2$ $>$ $K_1$, the convolution with kernel $\boldsymbol W_1 \in \mathbb{R}^{K_1 \times 1}$ and bias $b$ can be padded with 0 to get $\boldsymbol W_2 \in \mathbb{R}^{K_2 \times 1}$ without changing $b$. Thus, K$_1$ $\times$ 1 convolution can be converted to K$_2$ $\times$ 1 convolution.

\begin{figure}
  \centering
  \setlength{\abovecaptionskip}{0.1cm}
  \includegraphics[width=1.0\linewidth]{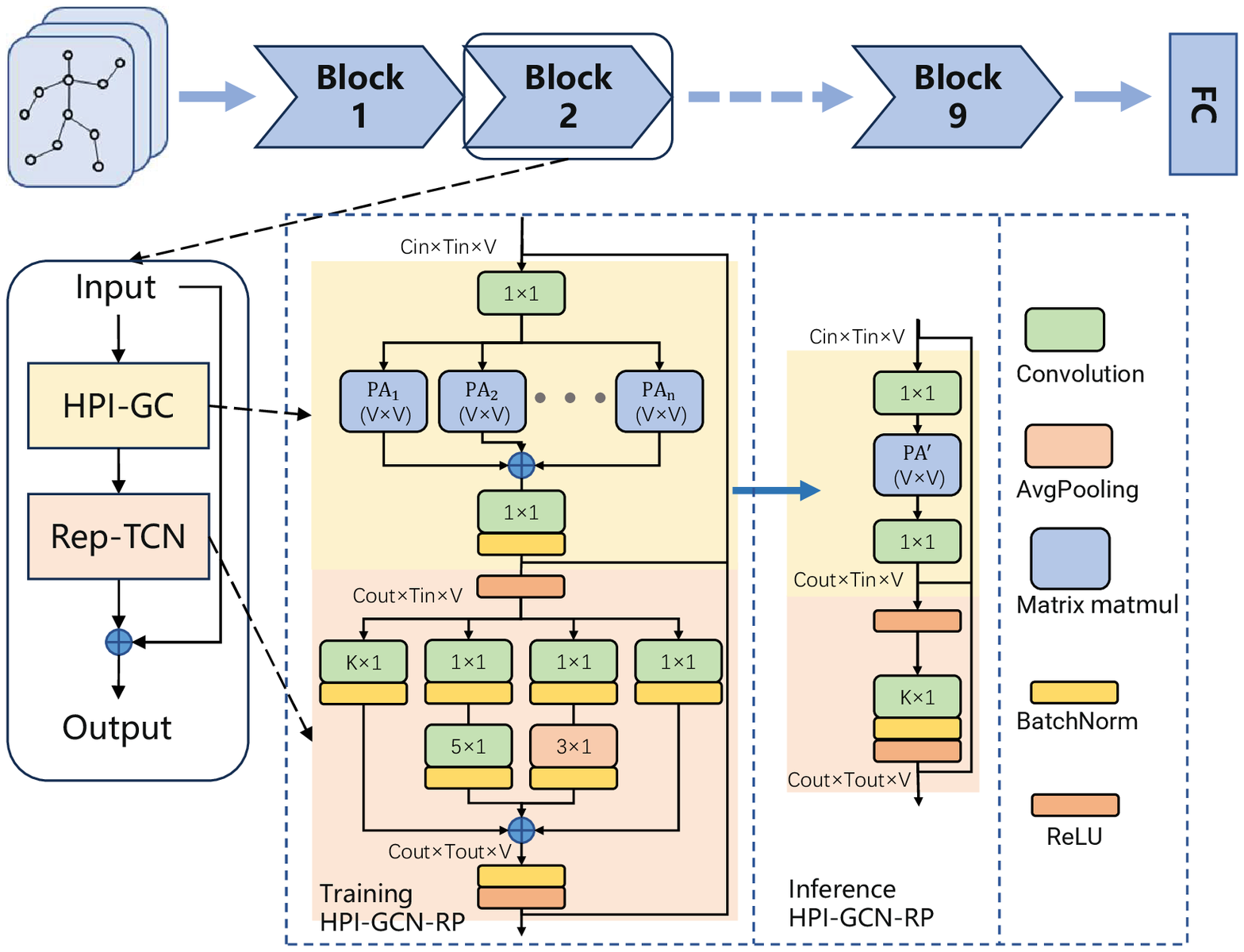}
  \caption{An overview of the HPI-GCN model. Specifically, it demonstrates the training and inference modes of the basic block of HPI-GCN-RP. We adopt a relatively complex training module, which can be transformed into a simple inference module through re-parameterization. Due to the linear transformation based on mathematical formulas, our inference module is equal to the training module, and it has fewer parameters, lower computational cost, and higher inference speed.
}
  \label{F1.2}
\end{figure}

\textbf{Blending 6:} For skeleton-based action analysis tasks, we designed Adjacency matrix multiplication fusion, where $\boldsymbol A \in \mathbb{R}^{V \times V}$ is a learnable matrix. The parallel multiple matrix multiplications output of an input $\boldsymbol X \in \mathbb{R}^{C \times T \times V}$ is given by
\begin{equation}
  \boldsymbol Y = \boldsymbol X * \boldsymbol A_1 + \boldsymbol X * \boldsymbol A_2.
\end{equation}
When the parameters of $\boldsymbol A$ are fixed in the inference process, it can be merged as
\begin{equation}
  \boldsymbol Y = \boldsymbol X * (\boldsymbol A_1 + \boldsymbol A_2) = \boldsymbol X * \boldsymbol A',
\end{equation}
Through Adjacency matrix multiplication fusion, the inference of multiple parallel matrix multiplications can be converted to a single matrix multiplication.

\subsection{Overall architecture}

We build two efficient models based on two HPI-GCs and Rep-TCN, named HPI-GCN-RP and HPI-GCN-OP, for skeletal-based human behavior analysis. Our model structure is similar to ST-GCN, consisting of 9 stacked basic blocks, and the block of HPI-GCN-RP is shown in Fig. \ref{F1.2}, with a training mode featuring a multi-branch learnable topology for improved feature extraction. Through re-parameterization techniques, the training mode can be linearly transformed into the inference mode, resulting in a simpler inference structure than ST-GCN and faster inference speed. The output channels of the block are [64, 64, 64, 128, 128, 128, 256, 256, 256]. Each block contains the HPI-GC module and Rep-TCN module. At the 4th and 7th blocks, the number of channels is doubled by HPI-GC, and the time dimension is halved by TCN. At the same time, in the joint stream, the input first passes through the Explicit Motion Modeling \cite{stgat} to increase the number of channels to 64, and in the bone stream, a 1 $\times$ 1 convolution is adopted to increase the number of channels to 64.

%-------------------------------------------------------------------------
\subsection{Re-Parameterized Temporal Convolution}

\begin{figure}
  \centering
  \setlength{\abovecaptionskip}{0.1cm}
  \includegraphics[width=0.7\linewidth]{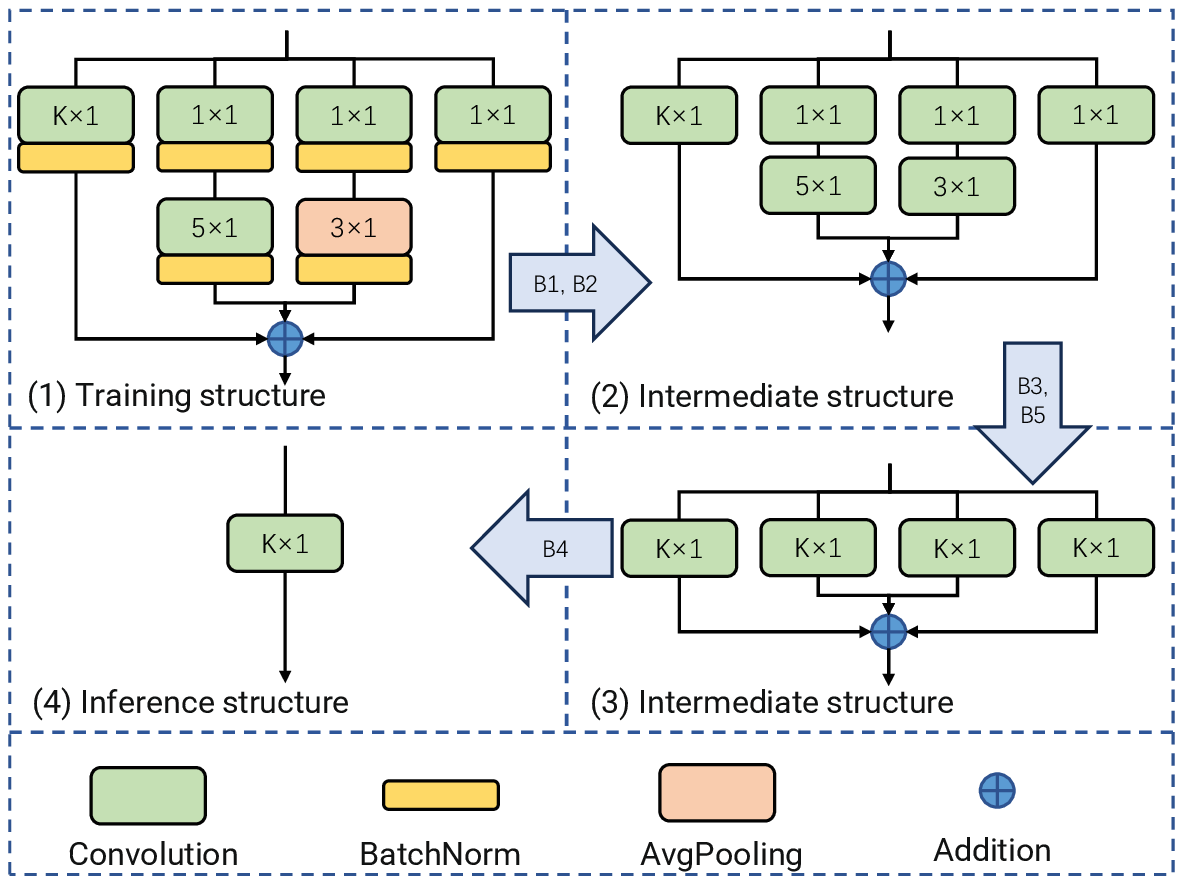}
 \caption{The Re-Parameterized Temporal Convolution Network (Rep-TCN). Fig. (1) is the training structure. Figs. (2) and (3) are the intermediate transformation process. Fig. (4) is the inference structure. B represents blending.}
  \label{F3}
\end{figure}

In the human behavior analysis task based on skeleton points, the ST-GCN \cite{stgcn} adopted TCN (the structure (4) in Fig. \ref{F3}) as training model to extract temporal features, which had a simple structure and fast inference speed but low accuracy. Most of the current SOTA methods \cite{ctrgcn,infogcn,hdgcn} utilize MS-TCN to extract temporal features, which adopts pooling and dilated convolution to extract multi-scale temporal information. Although this module exhibits good performance, it suffers from a complex multi-branch structure, resulting in a significantly slow inference speed. In summary, we propose Rep-TCN based on re-parameterization technique, which is trained with a multi-branch structure similar to MS-TCN (structure (1) in Fig. \ref{F3}) and transformed into an identical structure as TCN (structure (4) in Fig. \ref{F3}) for inference by re-parameterization. The Rep-TCN possesses both the high performance of MS-TCN and the swift inference speed of TCN.

The Rep-TCN could better capture the temporal relationships of the joints by modifying the convolution kernel size, layers and other hyperparameters. 
Moreover, other types of layers (such as pooling layers, residual connections) can be integrated into TCN. Different modules can be adopted for various tasks, but ensure no non-linear functions in the branches. Here, to align with MS-TCN, we opted for a 4-branch structure. Since the transformation involves re-parameterization, the difference in training models does not impact the inference structure. Through the transformations, our Rep-TCN can achieve satisfactory performance and inference speed simultaneously. 
The training with the structure (1) has higher accuracy than structure (4) because it extracts multi-scale features from the joints. After training, we utilize the re-parameterization method to convert the structure (1) to the structure (4) for inference. Compared with structure (1), the structure (4) has a simpler structure, higher inference speed, and no complex multi-branch structure, which makes it more friendly to modern GPU devices \cite{shufflenetv2}.

As shown in Fig. \ref{F3}, the transformation from the training structure (1) to the inference structure (4) consists of three steps: Firstly, blending 1 is applied to convert the 3$\times$1 average pooling into 3$\times$1 convolution, and then blending 2 is used to merge the convolution and batch normalization layer to obtain structure (2). Secondly, blending 3 is used to merge the second branch into 5$\times$1 convolution and the third branch into 3$\times$1 convolution, and then blending 5 is applied to convert all branches into K$\times$1 (K$ \geq $5) convolution to obtain structure (3). Thirdly, blending 4 is used to merge all branches into a single K$\times$1 convolution to obtain the inference structure (4). Since the transformation methods are mathematical transformations between parameters, the training structure (1) and inference structure (4) of Rep-TCN have the same accuracy.

%-------------------------------------------------------------------------
\subsection{High-Performance Inference GC}

\begin{figure*}
  \centering
  \setlength{\abovecaptionskip}{0.1cm}
  \includegraphics[width=0.9\linewidth]{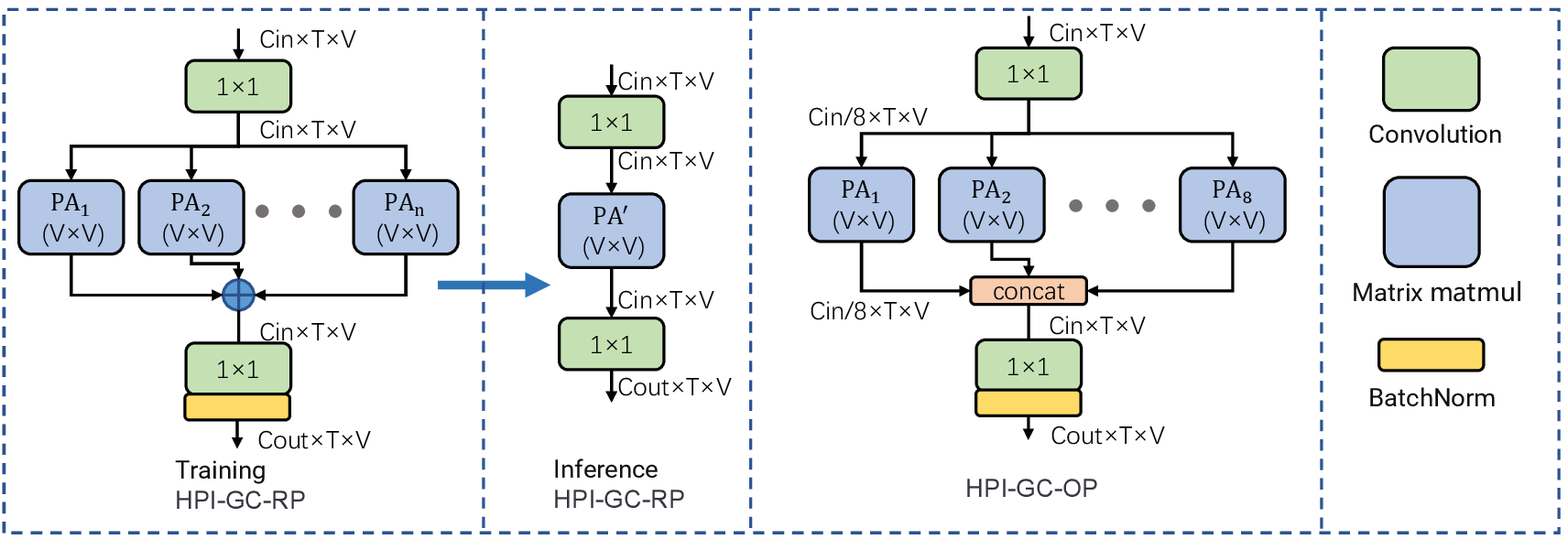}
  \caption{The two High-Performance Inference Graph Convolution (HPI-GC), where PA is a learnable adjacency matrix between the joints. Inference HPI-GC-RP is obtained from Training HPI-GC-RP with re-parameterization technique. HPI-GC-OP is derived from Inference HPI-GC-RP through the over-parameterization technique.
}
  \label{F4}
\end{figure*}

Fig. \ref{F4} shows two HPI-GC models that we designed. Based on the simple structure of ST-GCN, learnable adjacency matrices, and re-parameterization technique, we designed HPI-GC-RP. Firstly, a 1$\times$1 convolution is used for linear transformation, then some learnable Adjacency Matrix Parameters (PA) are applied for joints data modeling respectively. Finally, the 1$\times$1 convolution is used to transform the input channels into output channels.
By learning multiple adjacency matrices and integrating them into the inference network, our method provides more flexibility for the model and improves model representation to increase model accuracy. The training HPI-GC-RP is re-parameterized into Inference HPI-GC-RP through Blending 6, which reduces the number of inference branches without affecting the accuracy, and the transformation is as follows:
\begin{equation}
\begin{aligned}
  \boldsymbol Y &= \boldsymbol X * \boldsymbol {PA}_1 + \boldsymbol X * \boldsymbol {PA}_2 + ... + \boldsymbol X * \boldsymbol {PA}_n \\&= \boldsymbol X * \boldsymbol {PA}',
\end{aligned}
\end{equation}
where $\boldsymbol {PA}'=\boldsymbol {PA}_1+\boldsymbol {PA}_2 + ... + \boldsymbol {PA}_n$, $\boldsymbol {PA}\in \mathbb{R}^{V\times V}$, V represents the number of joint nodes. The structure of Inference HPI-GC-RP is simpler than ST-GCN, as its single-branch design has fewer branches compared to the three branches of ST-GCN. This simplicity makes it more suitable for large parallel devices like GPUs. Therefore, compared to ST-GCN, our HPI-GC not only delivers better performance but also achieves faster inference speeds.

The re-parameterization utilizes more training resources and computational costs, resulting in better performance and inference speed, but it cannot change the performance upper limit of the model. We enhance the performance upper limit of the model by introducing more nonlinear branches, a technique we refer to as over-parameterization. Specifically, we over-parameterize Inference HPI-GC-RP by increasing the number of nonlinear learnable matrices and designing HPI-GC-OP.
We divides the inputs into 8 groups according to channels and combines them with 8 learnable graph topology PA. By diversifying the matrix weights, more abstract information is extracted, which could enable the module to have stronger representation capabilities. The subsequent experiments have also demonstrated the stronger performance of HPI-GC-OP. Meanwhile, HPI-GC-OP has the same computational costs as Inference HPI-GC-RP, but due to the increased number of branches, it will increase a small amount of memory access and slightly decrease inference speed of the module.

%-------------------------------------------------------------------------
\section{Experiments}
In this section, we test our proposed HPI-GCN-RP and HPI-GCN-OP on two publicly skeleton-based action recognition datasets, and ablation studies are employed to validate the effectiveness and generality of our proposed re-parameterization and over-parameterization techniques. The computational speed is evaluated on an NVIDIA RTX 3090 GPU with a batch size of 64 and full precision (fp32), which is measured in milliseconds / iteration. The HPI-GCN-RP adopts 5 learnable PA.
%-------------------------------------------------------------------------

\subsection{Datasets}
\textbf{NTU-RGB+D 60.} The NTU-RGB+D 60 \cite{ntu60} is a large-scale human action recognition dataset consisting of 60 kinds of actions performed by 40 characters of different ages, which is captured by Microsoft's Kinect v2 with three cameras of different angles. The dataset is composed of 56,880 samples, and two benchmarks are recommended for evaluation: (1) Cross-Subject (X-Sub): The dataset is divided into 40,320 samples for training and 16,560 samples for testing based on the character ID; (2) Cross-Setup (X-View): The dataset is divided into 37,920 samples for training and 18,960 samples for testing based on the camera.
%This dataset is a valuable resource for deep learning research.

\textbf{NTU-RGB+D 120.} The NTU-RGB+D 120 \cite{ntu120} is an extended version of NTU-RGB+D 60, containing 120 categories, 114,480 action samples collected from 106 characters. The dataset is composed of 32 setups with different locations and backgrounds, and two benchmarks are recommended for evaluation: (1) Cross-subject (X-Sub): 53 subjects are used for training and the remaining 53 subjects for validation; (2) Cross-Setup (X-Set): Data with even setup ID is used for training and the rest of the data with odd ID for testing.

%-------------------------------------------------------------------------
\subsection{Implementation Details}
Our experiments are conducted on the NVIDIA RTX 3090 GPU based on PyTorch deep learning framework. The Nesterov momentum in SGD optimizer is 0.9 and weight decay is 0.0004. The CrossEntropyLoss is adopted in our experiment. We follow the training procedure of InfoGCN \cite{infogcn} with 110 epochs and the warm-up of the first 5 epochs. The learning rate is set to 0.1 and decays with a factor of 0.1 at epochs 90 and 100.
We set the dropout to 0.1 to prevent overfitting.
For the NTU-RGB+D and NTU-RGB+D 120 datasets, we follow the data pre-processing strategy of \cite{ctrgcn} by adjusting each sample to 64 frames and setting the batch size to 128.

%-------------------------------------------------------------------------

\subsection{Comparison with the State-of-the-Art}

\begin{table}
  \caption{\textbf{Comparisons of the top-1 accuracy (\%) and inference speed of state-of-the-art methods on the NTU-RGB+D 120 \cite{ntu120} (X-Sub) benchmark using multi-stream models.} Infer-speed represents the inference speed of single-stream models. The inference speed of multi-stream models is equal to the inference speed of single-stream models multiplied by the number of streams. We report the multi-stream model accuracy with the Multi-Modal Representation proposed by InfoGCN \cite{infogcn}.}
\begin{center}
\resizebox{\textwidth}{!}{
\begin{tabular}{l|cccc|c}
	\hline
	Model & \multicolumn{4}{|c|}{Acc(\%)} & Infer-speed \\
	 & Joint & 2-stream & 4-stream & 6-stream & (ms) \\ \hline
	HPI-GCN-RP    & 85.5    & 88.9    & 89.5    & 89.8    & \textbf{33}        \\
	HPI-GCN-OP    & \textbf{86.0}    & 89.0    & \textbf{89.8}    & 89.9    & 39        \\
	HPI-GCN-OP-K9  & 85.6    & \textbf{89.1}    & 89.7    & \textbf{90.1}    & 42        \\ \hline
	CTR-GCN \cite{ctrgcn}     & 84.9    & 88.7    & 88.9    & -       & 119       \\
	InfoGCN \cite{infogcn}    & 85.1    & 88.5    & 89.4    & 89.8    & 111       \\
	HD-GCN \cite{hdgcn}      & 85.7    & \textbf{89.1}    & \textbf{89.8}    & \textbf{90.1}    & 191       \\ \hline
\end{tabular}
}
\end{center}
  \label{Tab1}
\end{table}

\begin{table}
  \caption{\textbf{Comparisons of the top-1 accuracy (\%) against state-of-the-art methods on the NTU-RGB+D 120 \cite{ntu120} and NTU-RGB+D 60 \cite{ntu60} datasets at a similar inference speed.} Bold figures indicate the best value for each dataset. We give the accuracies based on two-stream, four-stream, and six-stream integration methods. The inference speed is measured with a batch size setting of 64 on the same NVIDIA RTX 3090 GPU.}
\begin{center}
\resizebox{\textwidth}{!}{
\begin{tabular}{l|c|c|c|c|c}
    \hline
    {Model}& \multicolumn{2}{|c|}{NTU-RGB+D 120} &\multicolumn{2}{|c|}{NTU-RGB+D 60}&Infer-speed\\
    &X-sub ($\%$)&X-set ($\%$)&X-sub ($\%$)&X-view ($\%$)&(ms)\\
    \hline
    ST-GCN \cite{stgcn} & - & - & 81.5 & 88.3 & -\\
    2s-AGCN \cite{2sagcn} (2s) & - & - & 88.5 & 95.1 & -\\
    Shift-GCN \cite{Shift-GCN} (4s) & 85.9 & 87.6 & 90.7 & 96.5 & - \\
    MS-G3D \cite{msg3d} (4s) & 86.9 & 88.4 & 91.5 & 96.2 & -\\
    PA-ResGCN-B19 \cite{song2020stronger} & 87.3 & 88.3 & 90.9 & 96.0 & -\\
    Dynamic GCN \cite{ye2020dynamic} (4s) & 87.3 & 88.6 & 91.5 & 96.0 & -\\
    MST-GCN \cite{mstgcn} (4s) & 87.5 & 88.8 & 91.5 & 96.6 & -\\
    EfficientGCN-B4 \cite{song2022constructing} & 88.3 & 89.1 & 91.7 & 95.7 & -\\
    
    %STF-Net \cite{STF-Net} &86.5 &88.2 &91.1 &96.5& -\\
    %SPIANet \cite{SPIANet} &89.2 &90.4 &92.8 &96.8& -\\
    %GSTLN \cite{GSTLN} &88.1 &89.3 &91.9 &96.6& -\\
    %RSA-Net \cite{RSANet} &88.8 &90.1 &92.7 &96.5& -\\
%LMA-GCN \cite{LMA-GCN} &83.2 &84.1 &88.2 &95.0& -\\
%FR-GCN  \cite{FR-GCN} (2s) &89.1 &90.2 &92.7 &96.7& -\\
%MADT-GCN \cite{MADT-GCN} (4s) &86.5 &88.2 &90.4 &96.5& -\\
LC-AGCN \cite{LC-AGCN} &84.2 &84.9 &89.3 &94.0 & -\\
IDGAN \cite{IDGAN}(4s) &88.9 &90.6 &92.5 &96.9& -\\

    STGAT \cite{stgat} (4s) & 89.4 & 90.8 & 93.0 & 97.4 & -\\
    CTR-GCN \cite{ctrgcn} (4s) & 88.9 & 90.6 & 92.4 & 96.8 & 476\\
    ST-GCN++ \cite{stgcnplus} (4s) & 88.6 & 90.8 & 92.6 & \textbf{97.4} &272\\
    InfoGCN \cite{infogcn} (4s) & 89.4 & 90.7 & 92.7 & 96.9 & 444\\
    InfoGCN \cite{infogcn} (6s) & 89.8 & 91.2 & 93.0 & 97.1 & 666\\

Mss-AGCN \cite{Mss-AGCN}(6s) &89.7 &91.0 &92.6 &96.8& -\\

    HD-GCN \cite{hdgcn} (2s) & 89.1 & 90.6 & 92.4 & 96.6 & 382\\
    HD-GCN \cite{hdgcn} (4s) & 89.8 & 91.2 & 93.0 & 97.0 & 764\\
    % HD-GCN\cite{hdgcn} (6s) & \textbf{90.1} & \textbf{91.6} & \textbf{93.4} & 97.2 & 1,146\\
    \hline

    %HPI-GCN-RP (2s) & 88.9 & - & 92.3 & 96.3 & 66 \\
    %HPI-GCN-RP (4s) & 89.5 & - & 92.6 & 96.6 &132 \\
    %HPI-GCN-RP (6s) & 89.8 & - & 92.8 & 96.8  &198 \\\hline

    HPI-GCN-OP-K9 (2s) & 89.1 & 90.3 & 92.4 & 96.5 & 84 \\
    HPI-GCN-OP-K9 (4s) & 89.7 & 91.1 & 92.8 & 96.9 &168 \\
    \textbf{HPI-GCN-OP-K9 (6s)} & \textbf{90.1} & \textbf{91.5} & \textbf{93.0} & 97.0  &\textbf{252} \\
    \hline
\end{tabular}
}
\end{center}

  \label{Tab2}
\end{table}

For a fair comparison, the recent state-of-the-art models have adopted multi-stream structures, and our multi-stream model adopts the same Multi-Modal Representation as InfoGCN. We compare our model with some of the most advanced methods in Tab. \ref{Tab1} on the NTU-RGB+D 120 X-Sub dataset. HPI-GCN-RP and HPI-GCN-OP adopt Rep-TCN with a maximum kernel size of 5, and HPI-GCN-OP-K9 adopts a maximum kernel size of 9 in Rep-TCN.

Through re-parameterization, HPI-GCN-RP achieves the fastest inference speed with satisfactory accuracy. In Tab. \ref{Tab1}, the single-stream model has an accuracy of 0.4\% higher than InfoGCN, and the inference speed of HPI-GCN-RP is 3.4 times faster.
HPI-GCN-OP adopt re-parameterization and over-parameterization techniques with a slight decrease in inference speed. The single-stream accuracy exceeds the current SOTA model HD-GCN by 0.3\%, and the inference speed is 4.9 times faster than HD-GCN.
Under the same time constraints, our HPI-GCN-OP can employ four-stream models to achieve 89.8\% performance, while HD-GCN can only employ single-stream model to achieve 85.7\% performance. Moreover, our six-stream accuracy also exceeds InfoGCN by 0.1\%, and the inference speed is 2.8 times faster than InfoGCN.
To further explore the effectiveness of re-parameterization and over-parameterization, we set the maximum kernel size of Rep-TCN in HPI-GCN-OP to 9.
HPI-GCN-OP-K9 achieves the same accuracy as the current SOTA model HD-GCN with six-stream, and the inference speed is 4.5 times faster, which demonstrates that our HPI-GCN-OP-K9 can balance both performance and inference speed simultaneously.
Overall, the experimental results demonstrate the effectiveness of our proposed re-parameterization and over-parameterization techniques.

In Tab. \ref{Tab2}, we report the comparison of our proposed HPI-GCN-OP-K9 with other SOTA models on NTU RGB+D 120 and NTU RGB+D 60. Although the main advantage of our model is the fast inference speed, our HPI-GCN-OP-K9 achieves competitive performance.
At the similar inference time, our six-stream HPI-GCN-OP-K9 not only has higher accuracy, but also 1.8 times faster than four-stream InfoGCN. Our six-stream HPI-GCN-OP-K9 has an inference speed 1.5 times faster than the two-stream HD-GCN and higher accuracy as well.

Within the same inference time constraint, our model achieves the best performance, which demonstrates the high efficiency and real-time nature. Notably, our approach is the first to adopt re-parameterization and over-parameterization techniques in GCNs, which has an excellent effect on skeleton-based human action recognition.

%-------------------------------------------------------------------------
\subsection{Ablation Study}
In this section, we analyze the effectiveness and generalizability of the proposed re-parameterization and over-parameterization techniques. All ablation experiments are conducted on the NTU RGB+D 120 (X-Sub) dataset in the joint stream.

\textbf{Re-parameterization.} The impact of our proposed re-parameterization technique on the model, as shown in Tab. \ref{Tab3}. The baseline is trained with the Inference HPI-GCN-RP in Fig. \ref{F1.2} without re-parameterization technique. The HPI-GCN-RP represents the training with the Training HPI-GCN-RP in Fig. \ref{F1.2}. The PA stands for the number of learnable Adjacency Matrix Parameters. The above models have the same inference structures, and the models trained with re-parameterization technique all perform better than the baseline. Among them, the accuracy steadily increases when the number of learnable PAs is increased. HPI-GCN-RP-PA5 with five learnable PAs achieves the highest accuracy, which demonstrates the effectiveness of re-parameterization technique.

\textbf{Over-parameterization.} We discuss the impact of our proposed over-parameterization technique on the model. As shown in Tab. \ref{Tab4}, on the basis of HPI-GCN-RP, HPI-GCN-OP employs over-parameterization technique to enhance the performance. The over-parameterization technique ensures that the inference computation of the model remains the same as the original model during the design phase with a slight increase in the number of parameters, which increases the model's capacity and expressiveness. Thus, the over-parameterized model can significantly improve the model performance with a small increase in the inference time, and HPI-GCN-OP boosts HPI-GCN-RP by 0.5\%, which demonstrates the effectiveness of over-parameterization technique.
\begin{table}
  \caption{\textbf{On NTU-RGB+D 120 \cite{ntu120} (X-Sub),} we gradually increase the number of learnable PAs in HPI-GCN-RP and report the accuracy and the inference speed.}
\begin{center}
\begin{tabular}{l|c|cc}
	\hline
	Model & Acc(\%) & Infer-speed(ms)  \\
	   \hline
	baseline          & 84.2  & 33             \\
	HPI-GCN-RP-PA1     & 85.1  & 33              \\
	HPI-GCN-RP-PA3     & 85.2  & 33               \\
	HPI-GCN-RP-PA5     & \textbf{85.5}  & 33       \\
 	\hline
\end{tabular}
\end{center}

  \label{Tab3}
\end{table}

\begin{table}
  \caption{\textbf{On NTU-RGB+D 120 \cite{ntu120} (X-Sub),} we report the accuracy, inference speed, computation and parameters of our proposed HPI-GCN-RP and HPI-GCN-OP.}
\begin{center}
\begin{tabular}{l|c|ccc}
	\hline
	Model & Acc & Speed & Flops & Parameters \\
	& (\%)& (ms) & (G) & (M) \\ \hline

	baseline     & 84.2  & 33 & 2.01     & 1.824          \\
	HPI-GCN-RP & 85.5  & 33 & 2.01     & 1.824          \\
	% GCN-OP  & -     & -  & -     & -          \\
	HPI-GCN-OP & \textbf{86.0} & 39 & 2.01 & 1.827  \\ \hline
\end{tabular}
\end{center}

  \label{Tab4}
\end{table}

\begin{table}
  \caption{\textbf{On NTU-RGB+D 120 \cite{ntu120} (X-Sub),} we report the re-parameterization technique to other SOTA models for accuracy tests on single-stream with the training hyperparameters as CTR-GCN \cite{ctrgcn}.}
\begin{center}
\begin{tabular}{l|c|c}
	\hline
	Model & Acc(\%) & Infer-speed(ms)  \\
	\hline

	CTR-GCN \cite{ctrgcn}     & 84.9  & 119              \\
	CTR-GCN-K5     & 84.8  & 97             \\
	CTR-GCN-RP-K5     & \textbf{85.3}  & 97               \\ \hline
	ST-GCN-K5 \cite{stgcn} & 83.4  & 48              \\
	ST-GCN-RP-K5  & \textbf{83.8}     & 48                \\ \hline
\end{tabular}
\end{center}
  \label{Tab5}
\end{table}

\textbf{Generalization.} To validate the generalization of our re-parameterization technique,
%To make a fair comparision, we keep the backbone of the SOTA models and only replace the temporal convolution.
we keep the backbone of the SOTA models and only replace the temporal convolution to make a fair comparison of the model performance and inference speed.
The experimental results are shown in Tab. \ref{Tab5}. Specifically, we construct CTR-GCN-K5 by replacing the MS-TCN in CTR-GCN with a 5$\times$1 temporal convolution (structure (4) in Fig. \ref{F3} with K=5). CTR-GCN-K5 performs 0.1\% lower than CTR-GCN, but the inference speed is 1.2 times faster than CTR-GCN. Then, we construct CTR-GCN-RP-K5 by applying the re-parameterization technique to CTR-GCN-K5, which is trained with structure (1) in Fig. \ref{F3} with K=5. CTR-GCN-RP-K5 is same as CTR-GCN-K5 in inference mode and achieves 0.5\% higher accuracy.
We further demonstrate the generalization of our Rep-TCN on ST-GCN.
For ST-GCN-K5, we retrain the model with the same hyperparameters as CTR-GCN to update the model performance, while applying the same training strategy to ST-GCN-RP-K5. ST-GCN-RP-K5 achieves 0.4\% higher accuracy while keeping the same inference speed. This demonstrates the generalization of Rep-TCN, which can improve the accuracy and inference speed of the current models.

In general, the models trained with re-parameterization technique achieve higher performance than the ones without adopting this technique. Moreover, the re-parameterization can further accelerate the inference speed, and the higher performance is achieved with our proposed over-parameterization technique.

\section{Applications and Limitations}
\label{Section5}
In the field of skeleton-based action analysis, our proposed HPI-GCN demonstrates outstanding performance at minimal cost. It holds great potential for a wide range of applications in scenarios with high real-time requirements, such as surveillance systems, virtual reality, and intelligent transportation. The flexibility of the model's training structure allows for dynamic adjustments based on diverse task requirements. For instance, in time-sensitive tasks, additional REP-TCN branches can be incorporated, while spatially sensitive tasks may benefit from a more complex spatial HPI-GC structure to extract richer information.
Our REP-TCN can directly replace the temporal convolution component in current SOTA models, significantly reducing model inference time while maintaining excellent performance, enhancing the practical utility of the model. Additionally, our re-parameterization graph convolution method serves as a specialized training technique, contributing to performance improvement.

Currently, our proposed re-parameterization graph convolutional network can only handle fixed-dimensional adjacency matrices. In essence, it is suitable for tasks with a fixed number of joints, such as skeleton-based action analysis. In the future, we plan to extend this approach to accommodate graph networks with a variable number of joints and implement re-parameterization to address dynamic changes in joint configurations.

%------------------------------------------------------------------------
\section{Conclusion}
In this work, we introduce re-parameterization into skeleton-based action analysis for the first time and design HPI-GCN-RP. Compared to ST-GCN, it achieves 1.5 times faster inference speed and 2.1\% performance improvement. Furthermore, we incorporate over-parameterization techniques into HPI-GCN-RP, resulting in HPI-GCN-OP, which achieves comparable performance to the current SOTA while boasting an inference speed five times faster. Finally, by replacing the temporal convolution block in current models with our proposed Rep-TCN, it simultaneously enhances both model performance and inference speed, demonstrating its effectiveness and generalization. 

\section*{Acknowledgements}
This work was supported by the Guangdong Basic and Applied Basic Research Foundation (2022A1515140126, 2023A1515011172), the Young and Middle-aged Science and
Technology Innovation Talent of Shenyang (RC220485), and The Royal Society Research Grant (RGS\textbackslash R2\textbackslash 222216)

\section*{Declarations}
The authors declare that they have no known competing financial interests or personal relationships that could have appeared to influence the work reported in this paper.



\bibliographystyle{elsarticle-num} 
\bibliography{egbib}% common bib file

\begin{thebibliography}{10}
\expandafter\ifx\csname url\endcsname\relax
  \def\url#1{\texttt{#1}}\fi
\expandafter\ifx\csname urlprefix\endcsname\relax\def\urlprefix{URL }\fi
\expandafter\ifx\csname href\endcsname\relax
  \def\href#1#2{#2} \def\path#1{#1}\fi

\bibitem{MADT-GCN}
Y.~Xia, Q.~Gao, W.~Wu, Y.~Cao, Skeleton-based action recognition based on
  multidimensional adaptive dynamic temporal graph convolutional network,
  Engineering Applications of Artificial Intelligence 127 (2024) 107210.

\bibitem{YUE2022287}
R.~Yue, Z.~Tian, S.~Du, Action recognition based on rgb and skeleton data sets:
  A survey, Neurocomputing 512 (2022) 287--306.

\bibitem{stgcn}
S.~Yan, Y.~Xiong, D.~Lin, Spatial temporal graph convolutional networks for
  skeleton-based action recognition, in: Thirty-Second AAAI Conference on
  Artificial Intelligence, 2018, pp. 7444--7452.

\bibitem{asgcn}
M.~Li, S.~Chen, X.~Chen, Y.~Zhang, Y.~Wang, Q.~Tian, Actional-structural graph
  convolutional networks for skeleton-based action recognition, in: Proceedings
  of the IEEE/CVF Conference on Computer Vision and Pattern Recognition, 2019,
  pp. 3595--3603.

\bibitem{2sagcn}
L.~Shi, Y.~Zhang, J.~Cheng, H.~Lu, Two-stream adaptive graph convolutional
  networks for skeleton-based action recognition, in: Proceedings of the
  IEEE/CVF Conference on Computer Vision and Pattern Recognition, 2019, pp.
  12026--12035.

\bibitem{song2022constructing}
Y.-F. Song, Z.~Zhang, C.~Shan, L.~Wang, Constructing stronger and faster
  baselines for skeleton-based action recognition, IEEE Transactions on Pattern
  Analysis and Machine Intelligence 45~(2) (2023) 1474--1488.

\bibitem{stgcnplus}
H.~Duan, J.~Wang, K.~Chen, D.~Lin, Pyskl: Towards good practices for skeleton
  action recognition, in: Proceedings of the 30th ACM International Conference
  on Multimedia, 2022, pp. 7351--7354.

\bibitem{msg3d}
Z.~Liu, H.~Zhang, Z.~Chen, Z.~Wang, W.~Ouyang, Disentangling and unifying graph
  convolutions for skeleton-based action recognition, in: Proceedings of the
  IEEE/CVF Conference on Computer Vision and Pattern Recognition, 2020, pp.
  143--152.

\bibitem{ctrgcn}
Y.~Chen, Z.~Zhang, C.~Yuan, B.~Li, Y.~Deng, W.~Hu, Channel-wise topology
  refinement graph convolution for skeleton-based action recognition, in:
  Proceedings of the IEEE/CVF International Conference on Computer Vision,
  2021, pp. 13359--13368.

\bibitem{infogcn}
H.~Chi, M.~H. Ha, S.~Chi, S.~W. Lee, Q.~Huang, K.~Ramani, Infogcn:
  Representation learning for human skeleton-based action recognition, in:
  Proceedings of the IEEE/CVF Conference on Computer Vision and Pattern
  Recognition, 2022, pp. 20186--20196.

\bibitem{hdgcn}
J.~Lee, M.~Lee, D.~Lee, S.~Lee, Hierarchically decomposed graph convolutional
  networks for skeleton-based action recognition, 2023 IEEE/CVF International
  Conference on Computer Vision (ICCV) (2022) 10410--10419.

\bibitem{shufflenetv2}
N.~Ma, X.~Zhang, H.-T. Zheng, J.~Sun, Shufflenet v2: Practical guidelines for
  efficient cnn architecture design, in: Proceedings of the European Conference
  on Computer Vision (ECCV), 2018, pp. 116--131.

\bibitem{rep0}
J.~Cao, Y.~Li, M.~Sun, Y.~Chen, D.~Lischinski, D.~Cohen-Or, B.~Chen, C.~Tu,
  Do-conv: Depthwise over-parameterized convolutional layer, IEEE Transactions
  on Image Processing 31 (2022) 3726--3736.

\bibitem{rep1}
S.~Guo, J.~M. Alvarez, M.~Salzmann, Expandnets: Linear over-parameterization to
  train compact convolutional networks, Advances in Neural Information
  Processing Systems 33 (2020) 1298--1310.

\bibitem{acnet}
X.~Ding, Y.~Guo, G.~Ding, J.~Han, {ACN}et: Strengthening the kernel skeletons
  for powerful cnn via asymmetric convolution blocks, in: Proceedings of the
  IEEE/CVF International Conference on Computer Vision, 2019, pp. 1911--1920.

\bibitem{repvgg}
X.~Ding, X.~Zhang, N.~Ma, J.~Han, G.~Ding, J.~Sun, Rep{VGG}: Making vgg-style
  convnets great again, in: Proceedings of the IEEE/CVF Conference on Computer
  Vision and Pattern Recognition, 2021, pp. 13733--13742.

\bibitem{dbb}
X.~Ding, X.~Zhang, J.~Han, G.~Ding, Diverse branch block: Building a
  convolution as an inception-like unit, in: Proceedings of the IEEE/CVF
  Conference on Computer Vision and Pattern Recognition, 2021, pp.
  10886--10895.

\bibitem{stgat}
L.~Hu, S.~Liu, W.~Feng, Skeleton-based action recognition with local dynamic
  spatial–temporal aggregation, Expert Systems with Applications 232 (2023)
  120683.

\bibitem{ntu60}
A.~Shahroudy, J.~Liu, T.-T. Ng, G.~Wang, {NTU RGB+D}: A large scale dataset for
  3d human activity analysis, in: Proceedings of the IEEE Conference on
  Computer Vision and Pattern Recognition, 2016, pp. 1010--1019.

\bibitem{ntu120}
J.~Liu, A.~Shahroudy, M.~Perez, G.~Wang, L.-Y. Duan, A.~C. Kot, {NTU RGB+D
  120}: A large-scale benchmark for 3d human activity understanding, IEEE
  Transactions on Pattern Analysis and Machine Intelligence 42~(10) (2019)
  2684--2701.

\bibitem{Shift-GCN}
K.~Cheng, Y.~Zhang, X.~He, W.~Chen, J.~Cheng, H.~Lu, Skeleton-based action
  recognition with shift graph convolutional network, in: Proceedings of the
  IEEE/CVF Conference on Computer Vision and Pattern Recognition, 2020, pp.
  183--192.

\bibitem{song2020stronger}
Y.-F. Song, Z.~Zhang, C.~Shan, L.~Wang, Stronger, faster and more explainable:
  A graph convolutional baseline for skeleton-based action recognition, in:
  proceedings of the 28th ACM International Conference on Multimedia, 2020, pp.
  1625--1633.

\bibitem{ye2020dynamic}
F.~Ye, S.~Pu, Q.~Zhong, C.~Li, D.~Xie, H.~Tang, Dynamic gcn: Context-enriched
  topology learning for skeleton-based action recognition, in: Proceedings of
  the 28th ACM International Conference on Multimedia, 2020, pp. 55--63.

\bibitem{mstgcn}
Z.~Chen, S.~Li, B.~Yang, Q.~Li, H.~Liu, Multi-scale spatial temporal graph
  convolutional network for skeleton-based action recognition, in: Proceedings
  of the AAAI Conference on Artificial Intelligence, Vol.~35, 2021, pp.
  1113--1122.

\bibitem{LC-AGCN}
K.~Wang, H.~Deng, Q.~Zhu, Lightweight channel-topology based adaptive graph
  convolutional network for skeleton-based action recognition, Neurocomputing
  560 (2023) 126830.

\bibitem{IDGAN}
J.~Huo, H.~Cai, Q.~Meng, Independent dual graph attention convolutional network
  for skeleton-based action recognition, Neurocomputing 583 (2024) 127496.

\bibitem{Mss-AGCN}
H.~Tian, Y.~Zhang, H.~Wu, X.~Ma, Y.~Li, Multi-scale sampling attention graph
  convolutional networks for skeleton-based action recognition, Neurocomputing
  597 (2024) 128086.

\end{thebibliography}

\end{document}